%% file: main.tex
\documentclass[letterpaper, 10 pt, journal, twoside]{template/IEEEtran}

\input{definitions}

\graphicspath{ {./media/} }

\begin{document}

\title{MotionSC: Data Set and Network for Real-Time Semantic Mapping in Dynamic Environments}

\markboth{IEEE Robotics and Automation Letters. Preprint Version. Accepted June, 2022}
{Wilson \MakeLowercase{\textit{et al.}}: MotionSC: Real-Time Semantic Mapping in Dynamic Environments} 

\author{Joey Wilson$^1$, Jingyu Song$^1$, Yuewei Fu$^1$, Arthur Zhang$^1$ \\
Andrew Capodieci$^2$, Paramsothy Jayakumar$^3$, Kira Barton$^1$, and Maani Ghaffari$^1$%
\thanks{Manuscript received: February 23, 2022; Revised May 20, 2022; Accepted June 20, 2022.}
\thanks{This paper was recommended for publication by Editor Cesar Cadena Lerma upon evaluation of the Associate Editor and Reviewers' comments.
Funding for M. Ghaffari was in part provided by NSF Award No. 2118818.} 
\thanks{DISTRIBUTION A. Approved for public release; distribution unlimited. OPSEC \#6152}
\thanks{$^1$J. Wilson, J. Song, Y. Fu, A. Zhang, K. Barton, and M. Ghaffari are with the University of Michigan, Ann Arbor, MI 48109, USA. {\texttt{\{wilsoniv,jingyuso,ywfu,arthurzh,bartonkl,maanigj\}\\@umich.edu}}}
\thanks{$^2$A. Capodieci is with Neya Systems Division, Applied Research Associates, Warrendale, PA 15086, USA. \texttt{acapodieci@neyarobotics.com}}
\thanks{$^3$P. Jayakumar is with the US Army DEVCOM Ground Vehicle Systems Center, Warren, MI 48397, USA. \texttt{paramsothy.jayakumar.civ@army.mil}}
\thanks{Digital Object Identifier (DOI): see top of this page.}
}


\maketitle

\begin{abstract}
This work addresses a gap in semantic scene completion (SSC) data by creating a novel outdoor data set with accurate and complete dynamic scenes. Our data set is formed from randomly sampled views of the world at each time step, which supervises generalizability to complete scenes without occlusions or traces. We create SSC baselines from state-of-the-art open source networks and construct a benchmark real-time dense local semantic mapping algorithm, MotionSC, by leveraging recent 3D deep learning architectures to enhance SSC with temporal information. Our network shows that the proposed data set can quantify and supervise accurate scene completion in the presence of dynamic objects, which can lead to the development of improved dynamic mapping algorithms. All software is available at \href{https://github.com/UMich-CURLY/3DMapping}{https://github.com/UMich-CURLY/3DMapping}. 
\end{abstract}

\begin{IEEEkeywords}
Data Sets for Robotic Vision, Deep Learning for Visual Perception, Mapping, Semantic Scene Understanding. 
\end{IEEEkeywords}

\IEEEpeerreviewmaketitle

\input{intro}
\input{preliminaries}

\input{method}
\input{results}
\input{conclusion}


{\footnotesize
\balance
\bibliographystyle{template/IEEEtran}
\bibliography{bib/strings-abrv,bib/ieee-abrv,bib/refs}
}

\end{document}

%% file: definitions.tex
\usepackage{graphicx}
\usepackage{caption}
\usepackage{amsmath,amssymb,enumerate}

\usepackage{amsthm}
\usepackage{mathtools}
\usepackage{breqn}
\usepackage[usenames,dvipsnames,svgnames,table, xcdraw]{xcolor}
\usepackage[colorlinks=true,pdfpagemode=UseNone,citecolor=black,linkcolor=black,urlcolor=BrickRed]{hyperref}
\usepackage{algorithm, algpseudocode}

\usepackage{blindtext}
\usepackage{gensymb}
\usepackage{xparse}
\usepackage{lipsum}
\usepackage{mathrsfs}
\usepackage[mathscr]{euscript}
\usepackage{times}
\usepackage[noadjust]{cite} 
\usepackage[numbers,sort&compress]{natbib}
\usepackage{multicol}
\usepackage[caption=false,font=normalsize]{subfig}
\usepackage{amsfonts}
\usepackage[utf8]{inputenc}
\usepackage[T1]{fontenc}
\usepackage{textcomp}
\usepackage{balance}
\usepackage{amsfonts}
\usepackage{soul}
\usepackage{multirow}
\usepackage{booktabs}
\usepackage{sidecap}
\usepackage{makecell}
\usepackage{array,float}

\renewcommand{\thefigure}{\arabic{figure}}

\definecolor{free}{rgb}{0.9608, 0.5882, 0.3922}
\definecolor{building}{rgb}{0.9608, 0.9020, 0.3922}
\definecolor{barrier}{rgb}{0.5882, 0.2353, 0.1176}
\definecolor{other}{rgb}{0.7059, 0.1176, 0.3137}
\definecolor{pedestrian}{rgb}{1, 0.3137, 0.3922}
\definecolor{pole}{rgb}{0.1176, 0.1176, 1}
\definecolor{road}{rgb}{0.7843, 0.1569, 1}
\definecolor{ground}{rgb}{0.3529, 0.1176, 0.5882}
\definecolor{sidewalk}{rgb}{1, 0, 1}
\definecolor{vegetation}{rgb}{1, 0.5882, 1}
\definecolor{vehicles}{rgb}{0.2941, 0, 0.2941}

\newcommand{\m}{\mathop{\mathrm{m}}}


%% file: intro.tex
\section{Introduction}

\IEEEPARstart{S}{cene} understanding in 3D is a keystone for mobile robotics. Scene understanding enables a robot to reason the environment and improve decision-making, internal navigation, path planning, and control execution. While 2D scene understanding is useful for robots, a 3D understanding ultimately provides an opportunity for more information, e.g., navigating under obstacles or understanding their size. Therefore, 3D scene understanding is necessary for autonomous vehicles \cite{SceneUnderstandingReview}, as the 2D information cannot support certain robotic applications due to the lack of depth information.

Scene understanding is generally broken into sub-tasks due to its broad scope, including object detection, scene categorization, semantic segmentation, depth estimation, tracking, and prediction~\cite{CVforAV}. These tasks are easier to conquer on their own, and have seen significant progress in recent years due to the development of 3D deep learning networks and rapid improvement of 3D sensors. The main challenges still facing 3D deep learning include memory and computation restrictions and working with sparse and unstructured data such as point clouds \cite{PointCloudReview}. Additionally, these sub-tasks are more straightforward to produce ground truth labels for than higher-level scene understanding tasks such as semantic scene completion, where a ground truth map of the world is nearly impossible to attain in outdoor, dynamic scenes \cite{SceneUnderstandingReview}.

\begin{figure}[t]
     \centering
    \subfloat[Ours]{
        \includegraphics[width=0.47\linewidth]{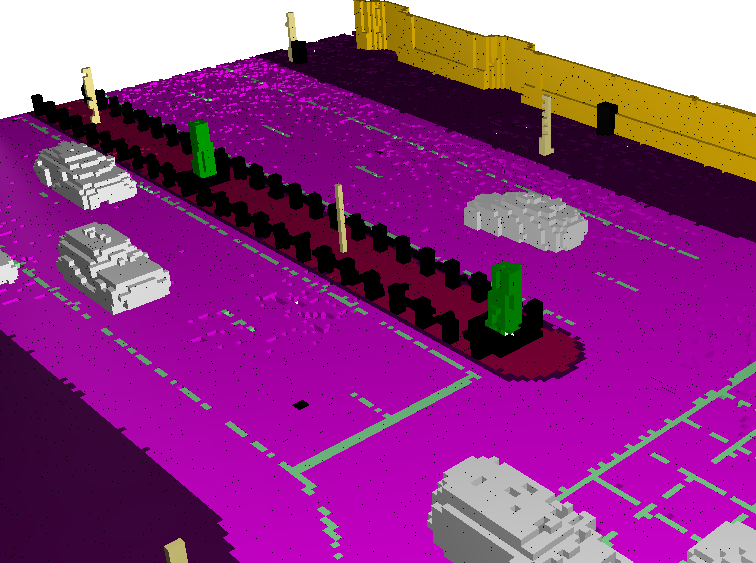}
        \label{fig:carla_image}
    }
    \subfloat[Semantic KITTI~\cite{KITTI}]{
        \includegraphics[width=0.47\linewidth]{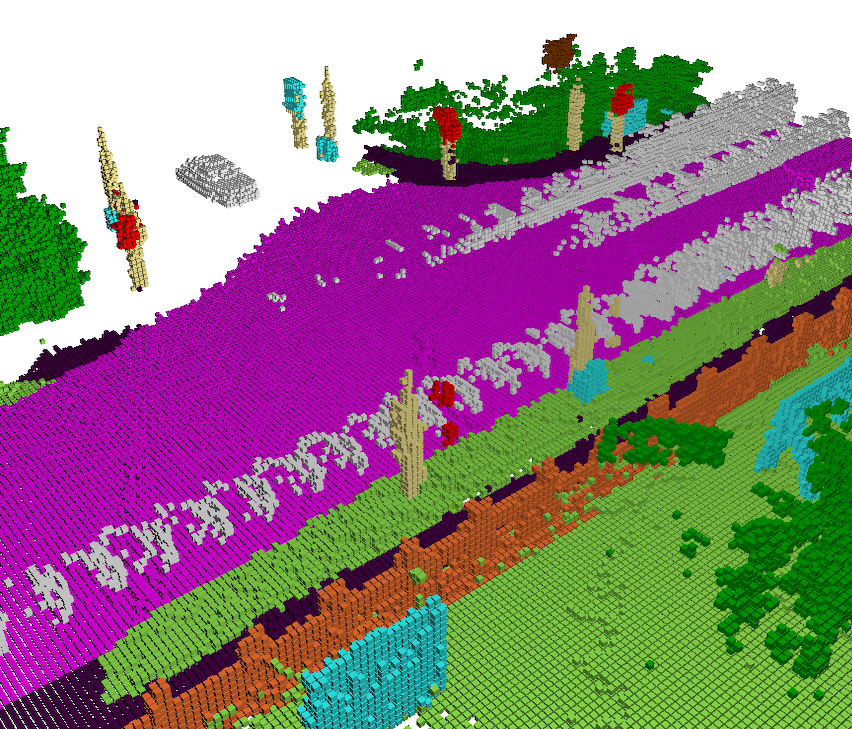}
        \label{fig:kitti_image}
    }
    \caption{Side-by-side comparison of our dynamic scene completion data set (left) with Semantic KITTI~\cite{KITTI} (right) on dynamic scenes from each respective data set. Moving vehicles, represented in white, leave traces in the KITTI data set yet are accurately and completely represented in our data set. Using multiple viewpoints, we reconstruct a dynamic scene without traces and with minimal occlusions.}
    \label{fig:CARLA better}
\end{figure}


Maps provide a unified framework to combine sensor information into a higher level of scene understanding with interpretable information. They store structured information which may be re-purposed later in tasks such as localization and path planning, where algorithms can retrieve locations of landmarks and driveable surfaces. This can provide benefits including increased reliability and predictability in diverse fields such as remote sensing, augmented reality, autonomous vehicles, and even medical devices. Although some research has proposed mapless navigation \cite{MaplessMP3, MaplessImitation, MaplessRL}, maps are still widely used due to their interpretability and success.  


Maps may include multiple layers of information such as traversability, semantic labels, and dynamic occupancy. Traversability maps generally construct a binary go-no-go map which may be used by autonomous vehicles for optimal path planning. Semantic maps include rich semantic labels for each cell, however generally only function in static scenes \cite{MappingSBKI, MappingROctoMap}. Dynamic occupancy maps construct binary labels for cells indicating free or occupied, and extend their domain to scenes with dynamic actors by incorporating scene dynamics \cite{MappingK3DOM, MappingDBKI, DynamicMapping}. While learning-based approaches have been attempted in 2D \cite{LearnMapMotioNet, LearnMapSelfSupervised}, most 3D maps rely on feature engineering which can decrease performance and efficiency.

Although mapping algorithms are generally hand-crafted, semantic scene completion (SSC) is a deep learning approach requiring supervision. In SSC, sensor data from a single scan is used to complete a dense local semantic scene
where completing the occluded portions of the scene may be accomplished by smoothing. 
However, the problem constraints leave time information unused. By adding temporal information, networks could gain insight into occlusions and dynamics and become more similar to supervised 3D semantic mapping algorithms.



The main challenges facing 3D semantic scene understanding are the memory and computation cost of 3D computer vision \cite{PointCloudReview, VoxelNet}, and the inability to obtain accurate ground truth data in the presence of dynamic objects and occlusion \cite{SSC_Survey}. To combat the exponentially increasing cost of 3D computer vision, some approaches include projecting to 2D views such as bird's-eye-view (BEV) \cite{LearnMapMotioNet, LMSCNet}, spherical-front-view (SFV) \cite{squeezeseg-spheri,rangeNet-spheri}, point pillars \cite{PointPillars}, or processing directly in 3D with point operations \cite{PointNet,PointNet++} or sparse convolutions \cite{second-sparse,MinkowskiNet}. Some more recent approaches leverage the speed of 2D convolutions while preserving height information by treating the vertical dimension of an occupancy grid as the channel~\cite{LearnMapMotioNet, LMSCNet}. This method maintains a high inference rate as fast as 100 Hz while achieving high performance in 3D object detection and SSC. 


A final challenge facing 3D maps is the availability of data sets. A complete scene is difficult to obtain in the real world due to dynamic objects. While indoor data sets 
\cite{indoor-nyuv2,indoor-matterport3d,indoor-scanNet,indoor-sun3d,SSCNet} may place multiple sensors throughout the scene to construct a 3D environment, this is not feasible for outdoor environments. Furthermore, currently existing synthetic data sets either contain only indoor environments, provide annotations for the scene only visible to the ego sensor, or remove dynamic objects altogether~\cite{SynthCity}. Therefore, mapping algorithms are commonly evaluated indirectly on object detection or motion prediction tasks. While data sets such as SemanticKITTI~\cite{KITTI} offer outdoor semantically labeled scenes, dynamic objects leave traces in the constructed map due to the inconsistency of the scene with respect to time. The lack of a complete outdoor 3D scene makes it impossible to directly learn or evaluate 3D mapping in dynamic, outdoor environments. 

In this paper, we propose to learn a real-time network that infers a dense ego-centric 3D semantic map by approaching SSC with temporal information. To quantify performance and supervise learning accurately, we construct a novel synthetic data set containing realistic outdoor driving scenes with complete semantic labels of free and occupied space. By randomly sampling multiple viewpoints of each scene, we ensure minimal occlusions and no traces left by dynamic objects. Our contributions are as follows.
\begin{enumerate}
    \item Develop a real-time deep neural network for 3D semantic mapping in dynamic environments.
    \item Demonstrate improvement over state-of-the-art semantic scene completion networks quantitatively and qualitatively. 
    \item Create a new data set for semantic scene completion and 3D mapping in dynamic outdoor environments to address the shortcomings in existing data sets. 
    \item Open source all data and network software for reproducibility and future developments at \href{https://github.com/UMich-CURLY/3DMapping}{https://github.com/UMich-CURLY/3DMapping}.
\end{enumerate}

%% file: preliminaries.tex
\section{Preliminaries}
\label{sec:preliminaries}
We first introduce current scene completion data sets, and modern approaches to the problem of semantic scene completion. We also mention trade-offs in storage methods of scenes, and introduce the motivation behind our choice in baselines.

\subsection{Data sets}
Current data sets cannot accurately measure scene completion or mapping performance in dynamic outdoor environments due to occlusions of moving objects. While an indoor scene may be captured from multiple viewpoints simultaneously, this is difficult to do in the real world~\cite{SSC_Survey}. Instead, existing data sets sample the ground truth map at time $t$ by aggregating data from sequential frames. However, this produces traces left behind by dynamic actors, leading to incorrectly labeled free or occupied space. An example of traces and occlusions in a modern scene completion data set is shown in Figure~\ref{fig:kitti_image}, where dynamic vehicles leave behind long white traces and parts of the scene are completely occluded. Current approaches solve this issue by avoiding dynamic objects altogether through rejection of dynamic objects during map generation or post-rejection of dynamic objects after the map is generated~\cite{EraseDynamic}. However, these convoluted methods are due to the unavailability of accurate dynamic scenes and are not ideal for obtaining a representation of the true, complete scene.

Existing outdoor scene completion data sets include SemanticKITTI~\cite{KITTI} and SemanticPOSS~\cite{SemanticPOSS}, or virtual data sets SynthCity~\cite{SynthCity} and more recently Paris-CARLA-3D~\cite{ParisCarla3D}. SemanticKITTI provides densely labeled voxelized scenes, but is susceptible to traces due to aggregating frames temporally, as shown in Figure \ref{fig:CARLA better}. While aggregating recent data to form a complete frame may aid with completeness, it violates the assumption that the data is independently and identically distributed (i.i.d.), as dynamic objects have moved between frames. \textcolor{black}{Although traces may be used to evaluate prediction of a vehicle's path, they ultimately hinder accurate evaluation as algorithms are penalized for correctly predicting free space.} SemanticPOSS also provides point clouds and semantic labels in dynamic scenes, yet is still susceptible to occlusions and traces for the same reason. SynthCity solves the issue of traces by removing dynamic objects at the cost of dynamic scenes. Finally, Paris-CARLA-3D provides a mixed synthetic and real data set using a mobile LiDAR sensor, but is still susceptible to traces for the same reason. Without a complete ground truth, it is impossible to correctly quantify performance on outdoor scene completion in dynamic environments, and difficult to supervise learning which properly handles dynamic objects. 

When creating a data set for SSC, there are a plethora of scene representations to consider, balancing resolution and information with data compression. Voxel grids are the most popular 3D representation of semantic scenes, however there exist several other less frequently used methods for efficiently storing 3D semantic information~\cite{SceneUnderstandingReview}. Point clouds are an efficient and direct method for storing scenes, which may be further optimized through meshes. However, both meshes and point clouds remain difficult representations for SSC, and few works have attempted this method due to challenging point generation. Additionally, sampling free space is more difficult as it is implicitly defined. Structured grids such as voxel grids or occupancy grids have the benefit of directly encoding semantic and free labels, and may be directly processed with convolutions. Despite high memory requirements due to explicitly defining free and occupied regions, their easy usage and interpretability make grids a frequent option. While grids may be implicitly encoded using a gradient field such as Signed Distance Function (SDF) or Truncated Signed Distance Function (TSDF), some SSC networks have found them to require a large computation time and add little benefit \cite{LMSCNet, TS3D, KITTI}. For real-time applications, this trade-off is not practical, and as a result voxel grids are commonly chosen.



\subsection{Semantic Scene Completion}\label{sec:ssc}
Scene Completion (SC) is a method of inferring the full geometry of a scene using either 2D or 3D observations. While research historically used interpolation methods, a subset of SC known as Semantic Scene Completion (SSC) has become popular due to advances in 3D deep learning networks. SSC differs from traditional SC by jointly inferring both semantics and geometry of the whole scene. Furthermore, the SSC task is made more difficult by the sparsity of the input data and incomplete ground truth, providing weakly guided supervision.


Semantic scene completion is a similar task to semantic mapping, however it is generally defined using sensor data from only a single frame. While there exist some deep learning 3D mapping methods, it is not a common task due to the lack of accurate outdoor dynamic data to supervise and quantify performance on~\cite{LearnDARNN, LearnMapSelfSupervised, MappingROctoMap, LearnRGBD}. SSC is currently a difficult task with minimal generalizability to real life due to the lack of accurately labeled dynamic scenes, as discussed in the previous section. 

There are several approaches to SSC which attempt to balance the complexities of 3D computer vision with a low run-time for practical purposes. 3D computer vision has become a more tractable problem due to innovative approaches such as PointNet~\cite{PointNet, PointNet++}, PointPillars~\cite{PointPillars}, VoxelNet~\cite{VoxelNet}, and sparse convolutions~\cite{MinkowskiNet, SparseConvNet}. These networks were able to reduce the memory and computation consumption of directly processing 3D data through viewing the problem in different ways. As a result, scene completion may be formulated in any of these formats through View-Volume networks, Volume networks, Hybrid networks, and Point-based networks~\cite{SSC_Survey}. 

Volume networks consist of 3D convolution to directly accomplish scene completion due to high levels of success in semantic segmentation, and the ability to aggregate contextual information from different levels~\cite{JS3CNet, Volume2}. This approach is limited by the large memory necessitated by directly processing 3D data, so is generally attempted with sparse convolutions such as in JSC3Net~\cite{JS3CNet}. View-Volume networks such as LMSCNet \cite{LMSCNet} \textcolor{black}{and SSCNet \cite{SSCNet}} use 2D convolutions in conjunction with 3D convolutions due to the efficiency of 2D convolutions. One method is to view the vertical axis of the 3D scene as a channel dimension, thus enabling the scene to be processed with 2D convolutions. This approach has also been taken in other 3D deep learning tasks with demonstrable success, such as MotionNet~\cite{LearnMapMotioNet} in the task of 3D object detection. A segmentation head then lifts the features from 2D to 3D to complete semantic segmentation of the 3D scene. Point-based networks operate directly on points to prevent discretization, but few works have investigated this avenue~\cite{Point1}. Finally, Hybrid networks combine any of the aforementioned methods~\cite{KITTI, Hybrid1, S3CNet, TS3D, Hybrid2}. 

%% file: method.tex
\section{Method}
Our methodology is split into two sections: first we construct a complete outdoor dynamic data set, and next we train a model from our data set. 

\begin{figure}[t]
    \centering
    \subfloat{\includegraphics[width=0.33\columnwidth]{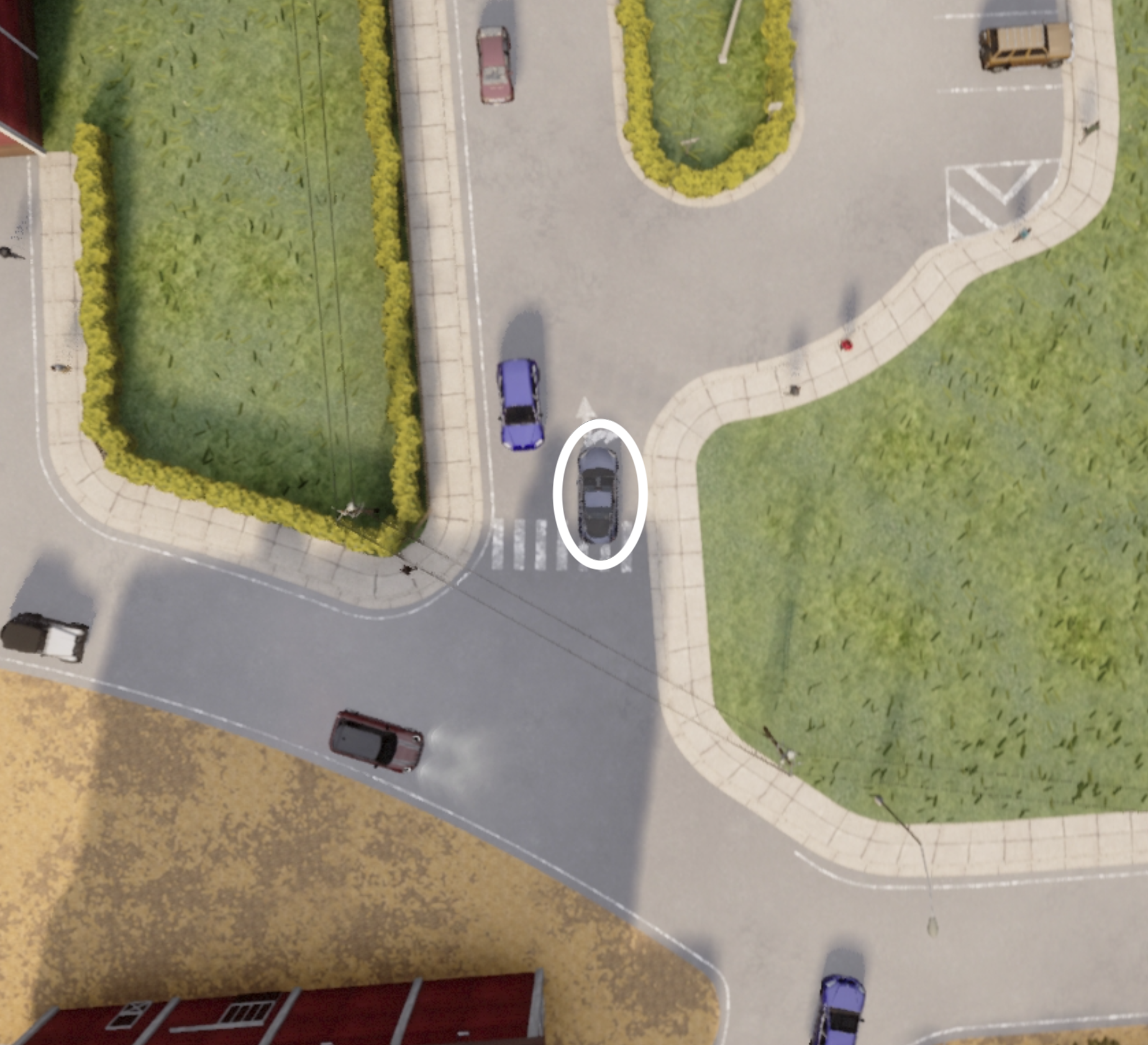}}
    \label{fig:bev_image}
    \subfloat{\includegraphics[width=0.315\columnwidth]{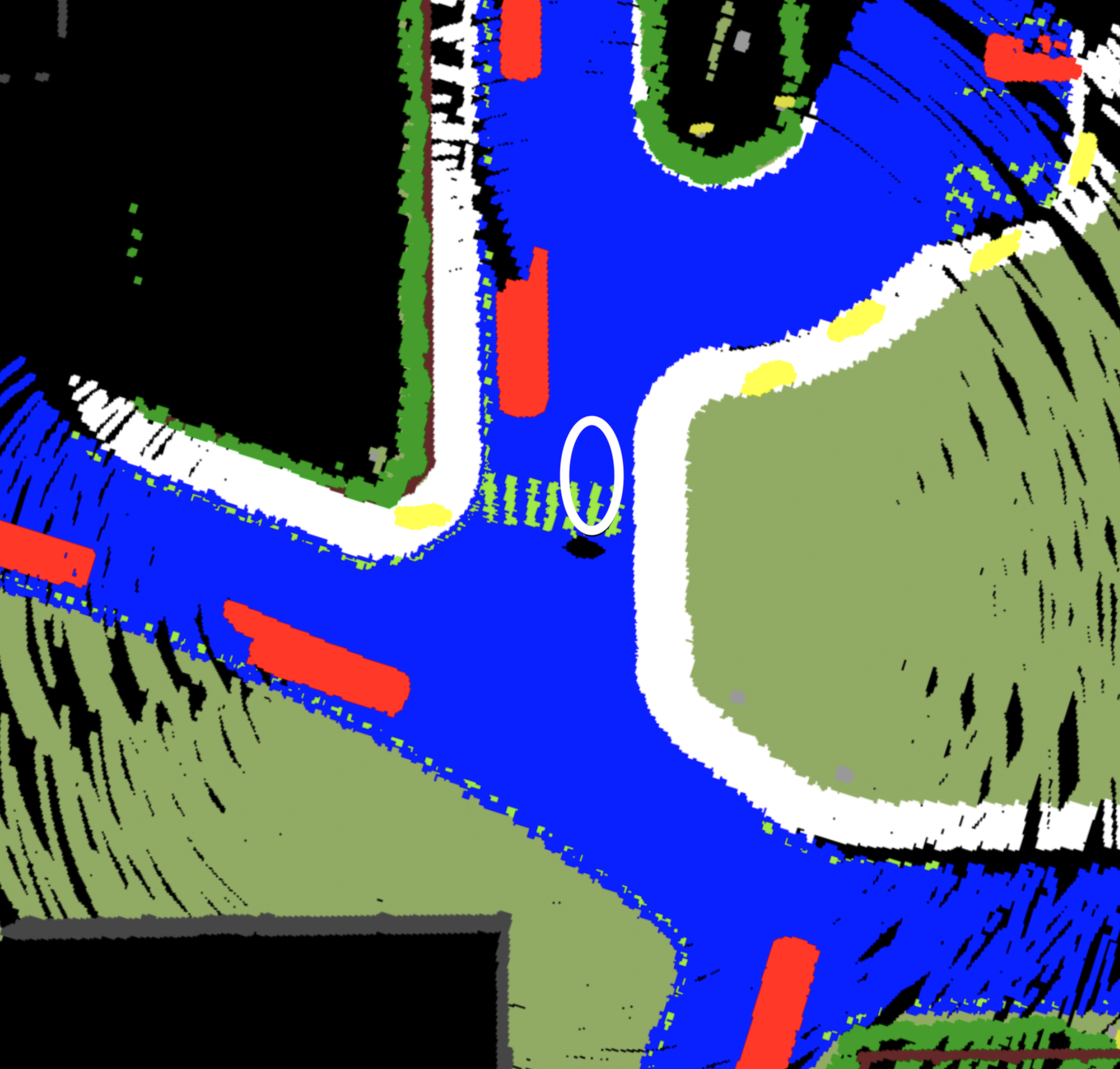}}
    \label{fig:concat_lidar}
    \subfloat{\includegraphics[width=0.315\columnwidth]{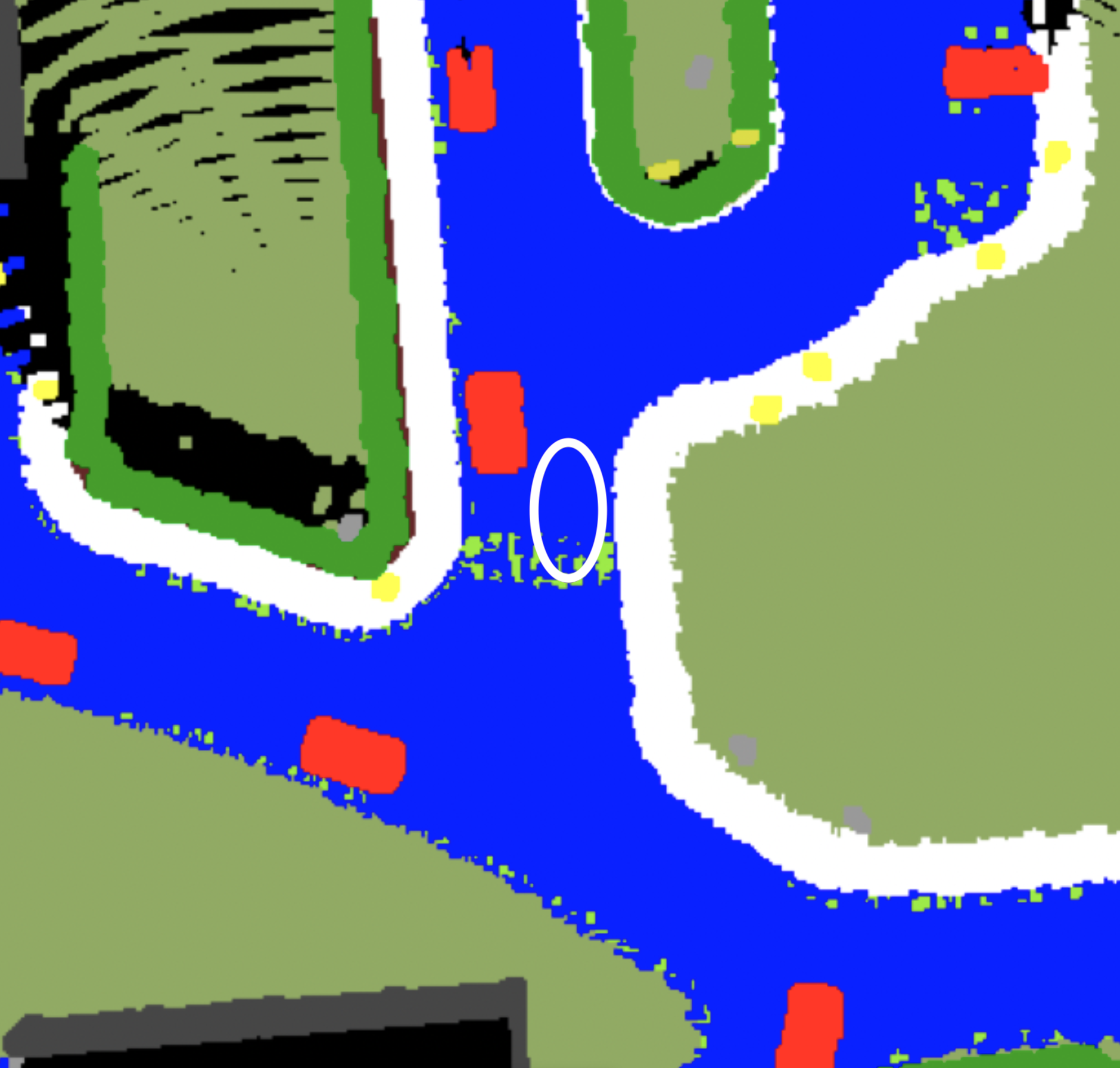}}
    \label{fig:all_lidar}
    \subfloat{\includegraphics[width=0.50\columnwidth]{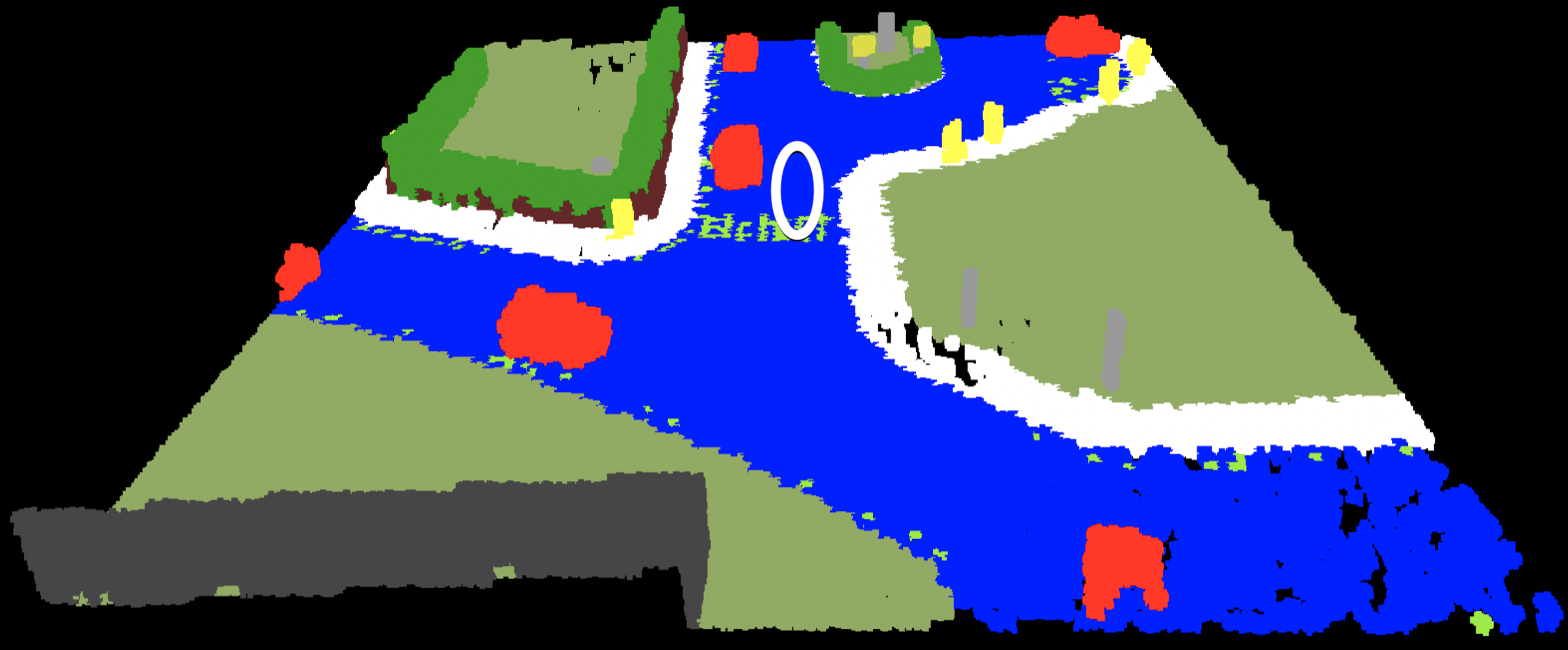}}
    \label{fig:cartesian_example}
    \subfloat{\includegraphics[width=0.485\columnwidth]{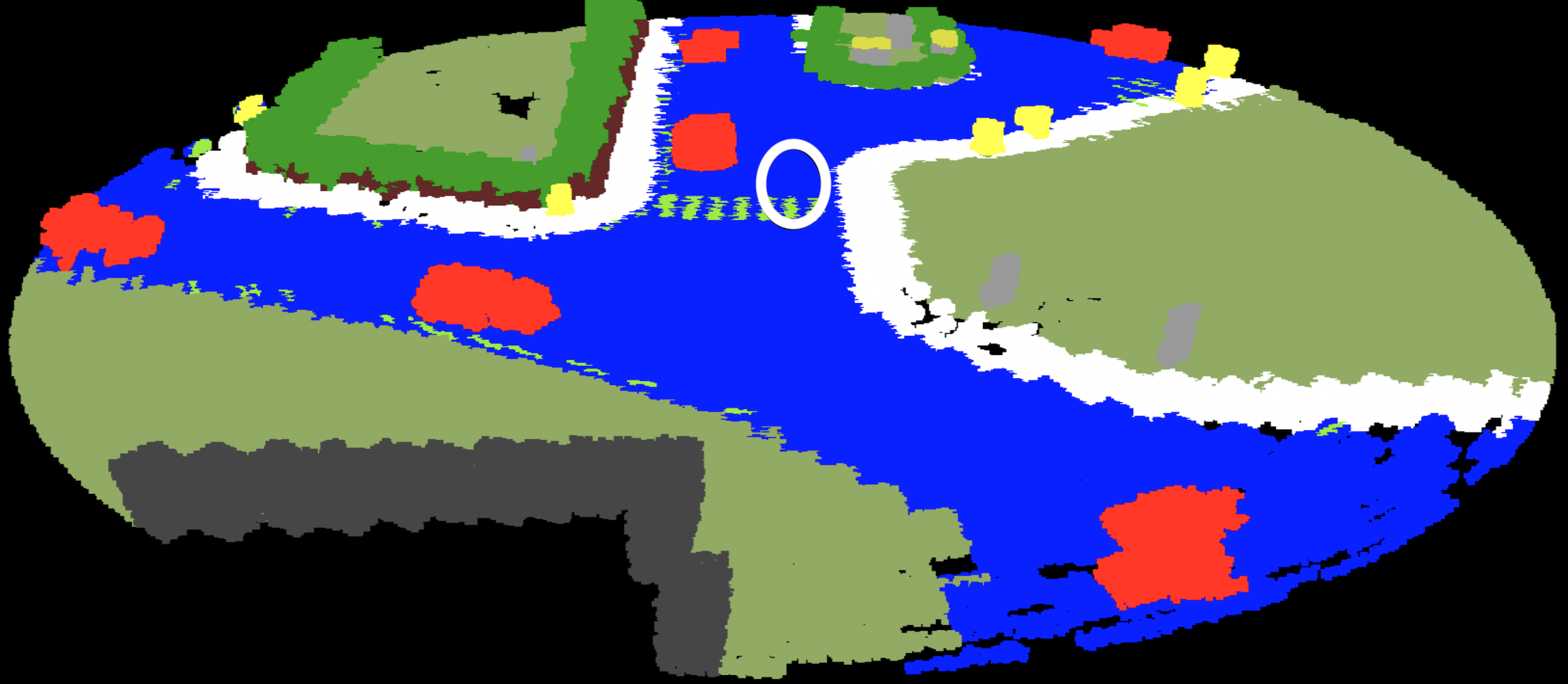}}
    \label{fig:cylindrical_example}
    \caption{An illustration of our approach to constructing a 3D dynamic outdoor driving data set. The ego vehicle is circled in all images, and is filtered out of the raw data. Top: The left image shows the Bird's Eye View (BEV) of a driving frame from our data set, the middle image demonstrates the current method of semantic scene generation, and the right image contains our method. Our method is able to accurately label free space without traces through the use of additional randomly placed sensors. Bottom: Constructed 3D semantic scene in Cartesian (Left) and Cylindrical (Right) coordinate systems. \textcolor{black}{While we do not use Cylindrical scenes in our evaluation, ground truth scenes may be found on our website.}} 
    \label{fig:RawData}
\end{figure}

\subsection{Data}\label{sec:MethodData}

We create a new outdoor driving semantic scene completion data set with accurate and complete scenes, which we call CarlaSC. Our data set is a large 3D synthetic outdoor driving scene completion data set with explicit free space and semantic labels, without the issues of occlusion and traces in Semantic KITTI~\cite{KITTI}. Ground truth semantic scenes are generated from multiple randomly placed LiDAR sensors in the CARLA~\cite{CARLA} simulator. While multiple cameras have been used for indoor static scenes, this is the first synthetic data set to extend the same approach to dynamic scenes. \textcolor{black}{Our data set is intended to correctly evaluate semantic mapping algorithms in dynamic scenes, and supervise training of dynamic algorithms. Although multiple view points may not be easily extended to the real world, we present ideas for how the static assumption of data collection may be removed to create accurate, complete scenes in Sections \ref{sec:MethodData} and \ref{sec:conclusion}}.

Our data set consists of 24 scenes in 8 dynamic maps with road users in various traffic settings.  We divide the maps into a train, validation, and test set where the train set contains the first six maps, the validation set contains map Town07, and the test set contains map Town10. Each map is used to create three scenes under low, medium and high traffic conditions, where each scene contains 3 minutes of data sampled at 10 Hz. Ground truth labels include point clouds with semantic segmentation and ego-compensated scene flow labels, ground truth pose and time labels, Bird's Eye View images, and the complete semantic scene. \textcolor{black}{Scenes are represented by 3D voxel grids due to explicit occupancy labels, and ease of use.}

\begin{figure}[t]
     \centering
     \includegraphics[width=0.8\columnwidth]{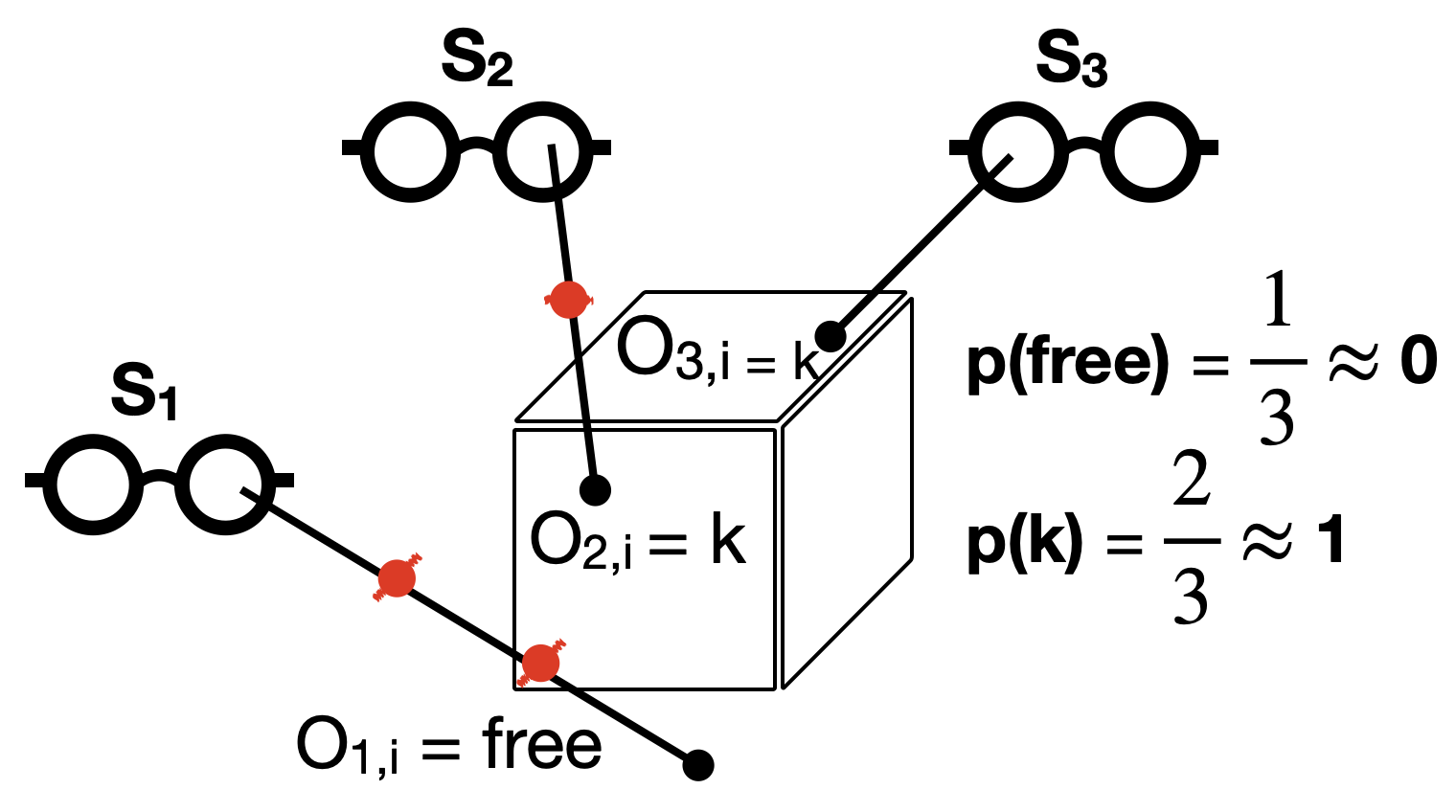}
    \caption{Toy illustration of how the semantic measurement for a single voxel $i$ is obtained.  \textcolor{black}{Multiple sensors, shown as eyeglasses, are uniformly distributed surrounding the ego vehicle and sample points with ground truth semantic labels from the environment.} Occupied points are shown in black, and free observations obtained from ray tracing are red. In this case, sensors 2 and 3 observe points in cell $i$ with label $k$. However, sensor 1 observes a point past the cell, resulting in a free space sample in cell $i$ from ray tracing. Therefore there are two $k$ measurements and one free measurement for voxel $i$, which is labeled as class $k$ after a majority vote.}
    \label{fig:SensorPlacement}
\end{figure}


We generate synthetic data from the CARLA simulator for its high definition, ability to generalize to the real world, and realism including autonomous pedestrians and vehicles, weather patterns, and traffic. Additionally, the CARLA simulator may be fully synchronized so that we may obtain ground truth pose and sensor data at every frame. While the CARLA API does not provide access to the mesh at each frame, we are able to semantically label the local volume around the ego vehicle by placing multiple sensors around the vehicle. 

Ground truth sensor data is obtained from simulated Velodyne 64-E LiDAR sensors, which are the same type used in KITTI \cite{KITTI}. The ego-sensor is placed on board the ego vehicle half a meter backwards from the center of the vehicle, and at the top of the vehicle 1.8 meters above the ground. The sensor samples at a frequency of 10 Hz with 130,000 points per scan created from 64 channels. The maximum range is 50 meters, and points which sample the ego vehicle or an invalid label from the simulator are discarded, creating a clean data set. Uniform noise is added to each point, such that each point is at most 2 centimeters from its true value. 


To create the semantic volume for each frame, we uniformly distribute 20 additional LiDAR sensors surrounding the ego vehicle, then aggregate their scans to obtain a complete scene. Any voxels without laser or free space measurements are rendered invalid, and excluded from evaluation. \textcolor{black}{The LiDAR sensors remain fixed relative to the ego-vehicle for the duration of the scene, and a new configuration is randomly generated for each driving sequence. While some data sets construct a ground truth scene in front of the ego vehicle, we construct a scene completely surrounding the vehicle due to the nature of LiDAR sensors.} By generating point clouds from multiple randomly placed sensors, we obtain voxelized labels more reflective of the true distribution, without the presence of traces and with minimal occlusions. We change the location of sensors between each scene to ensure that the whole data set is a better representation of the true data distribution, and to teach supervised networks to generalize to any view. Placing the sensors randomly assures that our data is i.i.d., which we can show guarantees convergence by the strong law of large numbers as the number of sensors and samples increases.

\begin{figure}[t]
\begin{algorithm}[H]
    \caption{Ray Tracing Algorithm}\label{alg:ray_tracing}
    \footnotesize
    \begin{algorithmic}[1]
    \State \textbf{Require:} semantic point cloud $P_s^t$, free space step $r$
    \State $O_s^t \gets \varnothing$ 
    \For{$x_i, y_i \in P_s^t$}
        \State $O_s^t \gets O_s^t \cup (x_i, y_i)$  \Comment{Occupied observation}
        \State $d \gets \lVert x_i \rVert - r$ 
        \State $\hat{x}_i \gets \frac{x_i}{\lVert x_i \rVert}$ 
        \While{$d > 0$} \Comment{Iterate towards sensor}
            \State $O_s^t \gets O_s^t \cup (d \cdot \hat{x}_i, \text{\emph{free}})$ \Comment{Free observation}
            \State $d \gets d - r$
        \EndWhile
    \EndFor
    \State $O_s^t \gets T_{0, s} O_s^t$ \Comment{Transform to ego-sensor frame}
    \State \textbf{Return:} sensor observations $O_s^t$
    \end{algorithmic}
\end{algorithm}
\end{figure}




Let $O^t$ indicate the set of observations of the world, $\mathcal{S}$ represent the set of $N$ randomly placed i.i.d. sensors $s$ sampled from a uniform distribution $U(\cdot)$ surrounding the ego vehicle, and $t$ be the current time. Then, we can write that: 


\begin{equation}\textcolor{black}{
    s \sim U( \begin{bmatrix}
    X_{min}, X_{max} \\
    Y_{min}, Y_{max} \\
    Z_{min}, Z_{max}
    \end{bmatrix}),
    }
\end{equation}

\textcolor{black}{
\begin{equation}
    p(O^t) = 
    \int_{\mathcal{S}}{p(O^t \mid s) p(s)} ds \:
    \approx \: \frac{1}{N}\sum_{i=1}^N p(O^t \mid s_i),
\end{equation}
where the set of observations is defined as
\begin{equation}\label{eq:combine sets}
    O^t = \{ O_s^t \mid \forall \, s \in \mathcal{S}\}.
\end{equation}
}

\textcolor{black}{This shows that our samples are representative of the expected observations, even in the presence of dynamic objects. Additionally, points from multiple uniformly distributed sensors may be aggregated into the ground truth scene without weighting as shown in \eqref{eq:combine sets}. While multiple viewpoints are difficult to achieve in the real world, the same approach of gathering i.i.d. samples may be applied through instance segmentation and reconstruction of non-deformable dynamic objects, which we have left as future work.} An example of the raw data produced for a single frame is shown in Fig.~\ref{fig:RawData}. As can be seen, there are no traces left by dynamic objects, and a more complete view of the scene is available, with fewer occlusions.

\begin{figure*}[t]
    \vspace{0.5mm}
     \centering
     \includegraphics[width=1.5\columnwidth]{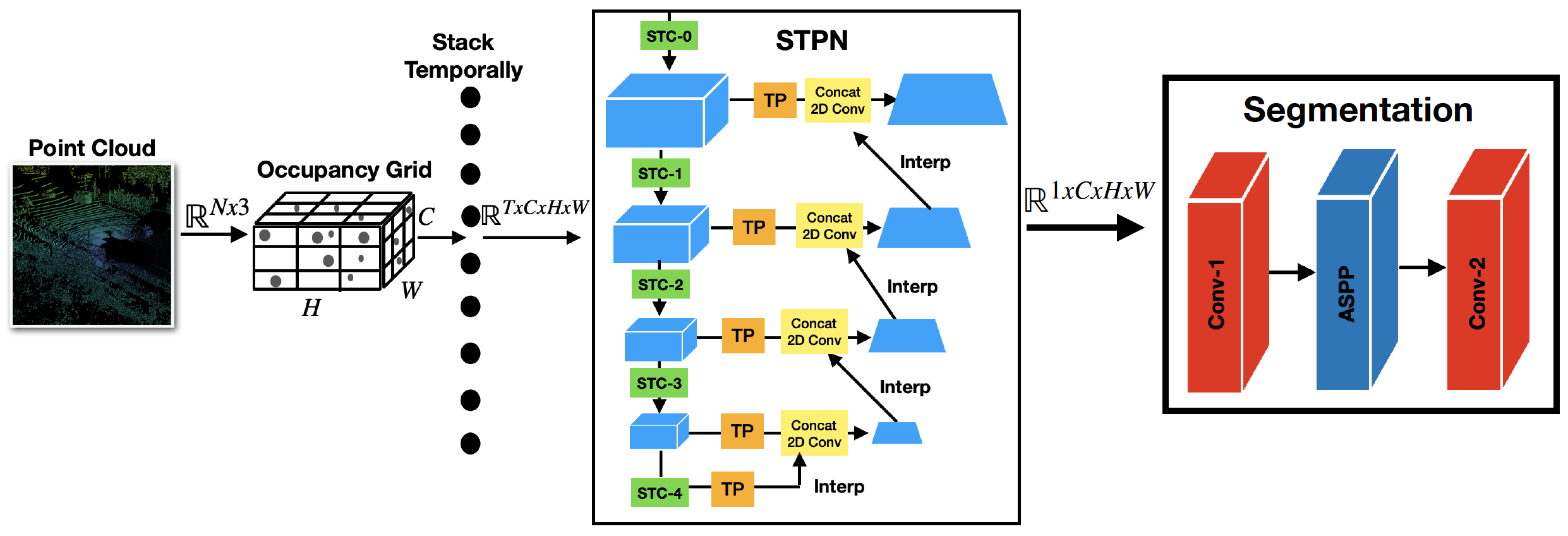}
    \caption{Proposed MotionSC network. \textcolor{black}{MotionSC is a view-volume scene completion network which incorporates a Spatio-Temporal Pyramid Network \cite{LearnMapMotioNet} backbone to incorporate temporal information in real-time. The input to the network is the past $T$ raw point clouds and corresponding poses, which are converted to a stack of the past $T \in \mathbb{N}$ binary occupancy grids.} For specific layer details, please see our website.}
    \label{fig:Proposed Network}
\end{figure*}


\textcolor{black}{To label free space in our 3D volume, we ray trace across each LiDAR ray in fixed distance increments, starting from the endpoint. Sampling in constant spacing ensures that free samples are also i.i.d.. An algorithm for ray tracing is shown in Algorithm \ref{alg:ray_tracing}, where the set of all occupied and free observations $O_s^t$  for sensor $s$ are computed from the point cloud $P_s^t$ observed by sensor $s$ in its frame of reference. Note that each point contains a geometric position $x \in \mathbb{R}^3$ and a semantic label $y \in \mathbb{N}$. $r$ is a hyper-parameter indicating the sampling distance between free space labels. Once all observations are computed, they are transformed to the on-board ego sensor using a transformation matrix where $s=0$ represents the ego-matrix. Finally, \eqref{eq:combine sets} shows how all observations are combined into one set $O^t$, which is used to compute the complete semantic scene. An illustration of the free space sampling and data aggregation method is shown in Fig.~\ref{fig:SensorPlacement}.}


\textcolor{black}{One advantage of our method is that it is scalable and open source, so that multiple resolutions or configurations may be generated. We publish two versions of the completed scenes on our website, one with the same dimensions and resolution of Semantic KITTI, and a slightly more coarse version. We choose the coarse scenes for our evaluations, where each semantically labeled scene is of the dimension (128, 128, 8), with a dimension of $-25.6$ to $25.6$ meters in the $x$ and $y$ axes, and $-2.0$ to $1.0$ meters in the vertical axis.} We set free space sampling interval $r=1.5 \m$ to avoid saturation of the data set from over-sampling and voxel discretization. For more information on how to use our data and the specifications, please see our website.


\begin{table}[b]
\centering
\scriptsize
\begin{tabular}{ p{2.5cm}|p{4cm} }
 \toprule
 \textbf{Remapped Class} & \textbf{Original Classes} \\
 \hline
 Other & Other, Sky, Bridge, Railtrack, Static, Dynamic, Water\\
 \hline
  Barrier & Fence, Wall, Guardrail\\
 \hline
 Poles & Pole, Traffic Light, Traffic Sign\\
 \hline
 Road & Road, Roadline\\
 \hline
 Ground & Ground, Terrain\\
 \bottomrule
\end{tabular}
\caption{Remapped classes and the original CARLA classes they contain.}
\label{table:ClassRemapping}
\end{table}

Our data set includes the raw CARLA labels obtained for every point, however we find that some classes are unobserved or exceedingly difficult to distinguish from one another by LiDAR sensor. \textcolor{black}{Therefore, we perform evaluation on a remapped version of the data, where semantic labels are mapped from the initial CARLA labels to a set of 11, as shown in Table \ref{table:ClassRemapping}.}


\begin{table*}[t]
\vspace{2mm}
\centering
\resizebox{0.75\textwidth}{!}{
\begin{tabular}{l|cc|ccccccccccc}
\multicolumn{1}{l}{\bf Method}&

\rotatebox{90}{\bf mIoU} &  
\rotatebox{90}{\bf Accuracy} &

\cellcolor{free}\rotatebox{90}{\color{white}Free} &
\cellcolor{building}\rotatebox{90}{\color{white}Building} &
\cellcolor{barrier}\rotatebox{90}{\color{white}Barrier} &
\cellcolor{other}\rotatebox{90}{\color{white}Other} & 
\cellcolor{pedestrian}\rotatebox{90}{\color{white}Pedestrian} & 
\cellcolor{pole}\rotatebox{90}{\color{white}Pole} &
\cellcolor{road}\rotatebox{90}{\color{white}Road} &
\cellcolor{ground}\rotatebox{90}{\color{white}Ground} &
\cellcolor{sidewalk}\rotatebox{90}{\color{white}Sidewalk} &
\cellcolor{vegetation}\rotatebox{90}{\color{white}Vegetation} &
\cellcolor{vehicles}\rotatebox{90}{\color{white}Vehicles} \\

\hline 

\vspace{-2mm} \\
LMSCNet SS~\cite{LMSCNet} & 42.53 & 94.64 & 97.41 & 25.61 & \textbf{3.35} & 11.31 & 33.76 & 43.54 & 85.96 & 21.15 & 52.64 & 39.99 & 53.09 \\
SSCNet Full~\cite{SSCNet, LMSCNet} & 41.91 & 94.11 & 96.02 & 27.04 & 1.82 & 13.65 & 29.69 & 27.02 & 88.45 & 25.89 & 65.36 & 33.29 & 52.78 \\
\bottomrule
\vspace{-2mm} \\
MotionSC (T=1) & 46.31 & 95.11 & 97.42 & 31.59 & 2.63 & 14.77 & 39.87 & 42.11 & 90.57 & 25.89 & 60.77 & \textbf{42.41} & \textbf{61.37} \\
MotionSC (T=5) & 45.35 & 95.00 & 97.43 & 29.48 & 2.54 & 17.48 & 41.87 & 43.43 & \textbf{90.90} & 22.08 & 58.43 & 35.79 & 59.41 \\
MotionSC (T=10) & 47.01 & 95.15 & 97.44 & 32.29 & 2.35 & 19.82 & 44.06 & \textbf{45.47} & 90.19 & 27.35 & 62.48 & 36.92 & 58.80 \\
MotionSC (T=16) & \textbf{47.45} & \textbf{95.57} & \textbf{97.60} & \textbf{34.91} & 2.66 & \textbf{22.86} & 37.78 & 43.87 & 90.12 & \textbf{28.31} & \textbf{66.20} & 41.59 & 56.08 \\
\end{tabular}
}
\caption{Semantic results on test set of CarlaSC. \textcolor{black}{The network is adjusted to accommodate smaller dimensions of $T$ by changing the temporal dimension of the Spatio-Temporal Convolution layer to $1$, effectively creating a fully-connected layer as the convolution kernel is dimension (1, 1, 1). The base model with $T=1$ uses 1,967 MB of memory, while MotionSC with a temporal stack of $T=16$ uses 2,427 MB of memory.}}
\label{tab:iou_reduced_classes}
\end{table*}

\subsection{MotionSC: Dense Local Mapping}
Next, we create a benchmark 3D dynamic semantic mapping neural network on our data set which we compare to baseline state-of-the-art scene completion algorithms. \textcolor{black}{The goal of our method, MotionSC, is to create a real-time dense local semantic mapping algorithm which may run on-board mobile robots in outdoor, dynamic scenes. Due to the memory and latency constraints of onboard applications, MotionSC is a view-volume scene completion network which primarily consists of 2D operations. MotionSC builds upon the real-time temporal backbone of MotionNet \cite{LearnMapMotioNet} which learns a latent space capable of reasoning about dynamic objects. A segmentation head then lifts the dimensionality of the feature space to predict a complete semantic scene. Our network synthesizes deep learning advancements in semantic scene completion and object detection to realize a higher level of scene understanding, and accomplish a new application of semantic mapping in dynamic environments. MotionSC is novel in that it combines a temporal backbone from object detection with a view-volume SSC network architecture to jointly track objects and complete semantic scenes.} A diagram of our network may be seen in Fig.~\ref{fig:Proposed Network}.


The input to MotionSC is a raw point cloud, which we convert to a 3D binary occupancy grid. We choose 3D occupancy grids instead of truncated signed distance function (TSDF) format since TSDF format has been found to require a greedy computation time with little benefit~\cite{LMSCNet, KITTI, TS3D}, which makes it impractical for a real-time dense local mapping algorithm. \textcolor{black}{Input point clouds are transformed to the current frame using pose information, as we found no correlation between performance and stack height $T$ if pose information was discarded.} 3D occupancy grids are created for each frame and maintained in a stack with a maximum depth of $T$ consecutive frames. By leveraging temporal information, we are able to train the network to reason about dynamic objects and occlusions. \textcolor{black}{The occupancy grids are a tensor of shape $(C, H, W)$ where the $H$ and $W$ represent the horizontal axes while $C$ represents the vertical axis, encoded as a channel dimension similar to a bird's eye view image.}

The temporally stacked occupancy grids are fed to a Spatio-Temporal Pyramid Network (STPN)~\cite{LearnMapMotioNet}, which treats the vertical axis as a channel dimension $C$ to process the four dimensional grid in real time using 2D convolutions and temporal pooling. The STPN is shown in Fig.~\ref{fig:Proposed Network}, and is a U-Net~\cite{UNet} structure composed of two key components: Temporal Pooling (TP) to aggregate information across frames, and Spatio-Temporal Convolution (STC) to increase feature size while reducing the temporal and spatial dimensions. \textcolor{black}{The output of the STPN is a feature vector of shape $(1, C, H, W)$, where temporal information is implicitly encoded within the channel dimension.} \textcolor{black}{TP is a consistent operation throughout all layers, and applies a max pooling operation over the 4D tensor across the temporal axis, reducing the $T$ dimension to size 1.} STC is a series of 2D convolutions dimensions to grow the channel size and reduce the spatial dimension by half, sometimes followed by a 3D convolution over the $(T, H, W)$ dimensions. \textcolor{black}{For $T<3$, temporal convolution with a width of $3$ may not be performed, so we resize the filter to shape $(1, 1, 1)$ which effectively performs a linear layer operation.} Upsampling deeper layers to combine with shallow layers is done by interpolation over the $(H, W)$ dimensions by a factor of 2, then concatenating with the next sequential layer and applying 2D convolutions. For more information on STPN see~\cite{LearnMapMotioNet}, and on our layer hyper-parameters see our website.

\textcolor{black}{A segmentation head lifts the dimensionality of the latent tensor from $(1, C, H, W)$ to $(K, C, H, W)$ in order to assign a semantic label to every 3D voxel. We use the segmentation head from LMSCNet \cite{LMSCNet}, as LMSCNet is also a view-volume scene completion network which has achieved high performance.} The segmentation head performs 3D convolutions to lift the feature dimension from 1 to $K$, and learns spatial relations through an Atrous Spatial Pyramid Pooling (ASPP) block \cite{ASPP1, ASPP2_improve}. For inference, a softmax layer is applied to obtain probabilities per class. 

%% file: results.tex
\section{Results and Discussion}
\label{sec:Results}
We establish baselines using state-of-the-art open source SSC networks on our data set, then compare our proposed real-time dynamic semantic mapping network, MotionSC, to establish a benchmark. We perform a temporal ablation study to demonstrate that temporal information enhances our network to achieve semantic and geometric completion results on par or better on both metrics. \textcolor{black}{We also include a comparison on the real-world Semantic KITTI data set to demonstrate that our method works successfully on real data. 
} Finally, we include a qualitative comparison from images on our test set of each model. 

\subsection{Quantitative Evaluation}
We perform a comparison of our method with several state-of-the-art SSC networks on the test set of our data set, and the real-world Semantic KITTI \cite{KITTI} data set. We choose to compare our network on our data set with LMSCNet \cite{LMSCNet} and SSCNet \cite{SSCNet}, as they are the best available open-source view-volume SSC networks at the time. \textcolor{black}{We use the single-state (SS) variation of LMSCNet and full variation of SSCNet, as they were shown to have the best performance in \cite{LMSCNet}. While these networks do not take advantage of temporal information, they serve as a strong baseline for MotionSC on our data set. To ensure a fair comparison, we perform an ablation study of the temporal stack height. At $T=1$, our network reduces to a view-volume SSC model, while at larger stack sizes it uses past information. As mentioned in Section \ref{sec:ssc}, there are few open-source supervised mapping algorithms for dynamic scenes, and as a result state-of-the-art SSC networks make competitive baselines.}

LMSCNet and SSCNet are view-volume networks with low inference times and memory usage. We imitate each network's learning parameters and process as closely as possible and choose the best model according to the mean intersection over union (mIoU) on the validation set. Models are trained for 24 to 48 hours, and model weights and training data are recorded and are available open-source for reference. We train the models until convergence is reached, which we found to be 10 epochs for MotionSC and 20 epochs for the baselines. 

\textcolor{black}{MotionSC is compared with the baselines on the test set of CarlaSC and Semantic KITTI. Concretely, the input to each algorithm is the past $T$ scans from the ego-vehicle LiDAR, and the output is a full scene containing a predicted semantic label for each voxel. Predictions are evaluated at every ground truth voxel containing laser measurements, including voxels occluded to the ego vehicle.} In line with \cite{KITTI}, we compute performance over geometric completeness and semantic completeness. 


\textcolor{black}{Table \ref{tab:iou_reduced_classes} shows the results of our model compared to the baselines on the test set, including an ablation study of the effect of $T$. Comparing MotionSC with $T=1$ with the baselines, we find that our model achieves both a higher semantic accuracy and mean intersection over union (mIoU). As the temporal stack height increases, MotionSC improves at nearly all metrics, most evident by the mIoU score. A stack height of $T=16$ corresponds to the last 1.6 seconds of information, and could be further increased at the cost of more memory.}
A similar pattern emerges when measuring performance on geometric completeness, shown in Table \ref{table:geometric_comparsion_sub}. Geometric completeness measures binary occupancy by mapping all non-free semantic classes to a single label. 
This metric indicates that MotionSC gains an advantage in completeness by leveraging temporal information to mitigate occlusions.


\renewcommand{\arraystretch}{1.2}

\begin{table}[t]
\vspace{2mm}
\centering
\scriptsize
\begin{tabular}{ l|c|c|c}
 \toprule
 \textbf{Method} & \textbf{Precision (\%)} & \textbf{Recall (\%)} & \textbf{IoU (\%)} \\
 \bottomrule
 \hline
 SSCNet Full & 85.87 & \textbf{93.05} & 80.69\\
 \hline
 LMSCNet SS & \textbf{95.62} & 89.5 & 85.98\\
 \bottomrule
 \hline
 MotionSC (T=1) & 93.32 & 92.16 & 86.46\\
 \bottomrule
 MotionSC (T=5) & 94.76 & 90.57 & 86.25 \\
 \bottomrule
 MotionSC (T=10) & 93.17 & 92.43 & 86.56 \\
 \bottomrule
 MotionSC (T=16) & 94.61 & 91.77 & \textbf{87.21} \\
 \bottomrule
\end{tabular}
\caption{Geometric Completeness on test set of CarlaSC. MotionSC has a better geometric IoU than the baselines at $T=1$, which is again increased as the temporal stack height grows. Although LMSCNet achieves the best precision, it has lower recall, meaning that it fails to identify a larger portion of the occupied space.}
\label{table:geometric_comparsion_sub}
\vspace{-3mm}
\end{table}

\begin{table}[t]
\centering
\scriptsize
\begin{tabular}{ l|c|c|c}
 \toprule
 \textbf{Method} & \textbf{Geom. IoU (\%)} & \textbf{Sem. mIoU (\%)} & \textbf{Latency (ms)} \\
 \bottomrule
 \hline
 SSCNet Full \cite{SSCNet} & 49.98 & 16.14 & 2.18\\
 \hline
 LMSCNet SS \cite{LMSCNet} & 56.72 & 17.62 & 4.86\\
 \hline
 \textbf{MotionSC (T=1)}& \textbf{56.85} & 18.42 & 5.72\\
 \hline
 JS3CNet \cite{JS3CNet} & 56.6 & \textbf{23.8} & 166.2\\
 \bottomrule
\end{tabular}
\caption{\textcolor{black}{Comparison of published methods on test set of Semantic KITTI \cite{KITTI} over Geometric IoU, Semantic mIoU, and latency metrics. We only compare our method with T=1 due to the restrictions of the scene completion competition. Latency is measured on an NVIDIA GeForce RTX 3090 GPU, averaged over one hundred repetitions.}}
\label{table:kitti_comparison}
\vspace{-4mm}
\end{table}

\textcolor{black}{Finally, we quantitatively evaluate our model on the test set of Semantic KITTI, shown in Table \ref{table:kitti_comparison}. For this comparison, we use $T=1$ as MotionSC reduces to a view-volume semantic scene completion network. On Semantic KITTI, MotionSC again outperforms SSCNet Full and LMSCNet SS, demonstrating that the performance patterns transfer from our data set to the real world. We include an additional comparison with the volume network JS3CNet \cite{JS3CNet}. JS3CNet achieves greater semantic performance than the baselines and our model, however it has lower geometric completeness than MotionSC and a slower inference rate due to costly 3D convolutions.} Note that incorporating temporal information should logically see the same gains in geometric and semantic completeness, thereby improving MotionSC in real world robotic applications.

\subsection{Qualitative Evaluation}
We compare MotionSC qualitatively with SSCNet Full and LMSCNet SS on the test set of our data set by generating predictions for each model and comparing the predictions visually in a side-by-side comparison. \textcolor{black}{We also include illustrations of MotionSC when $T=1$ and $T=16$ to demonstrate the effect of temporal information. We choose a scene with an occlusion, as temporal information allows the network to complete regions which are currently occluded.}

\begin{figure}[ht]
    \vspace{0.5mm}
     \centering
     \includegraphics[width=\columnwidth]{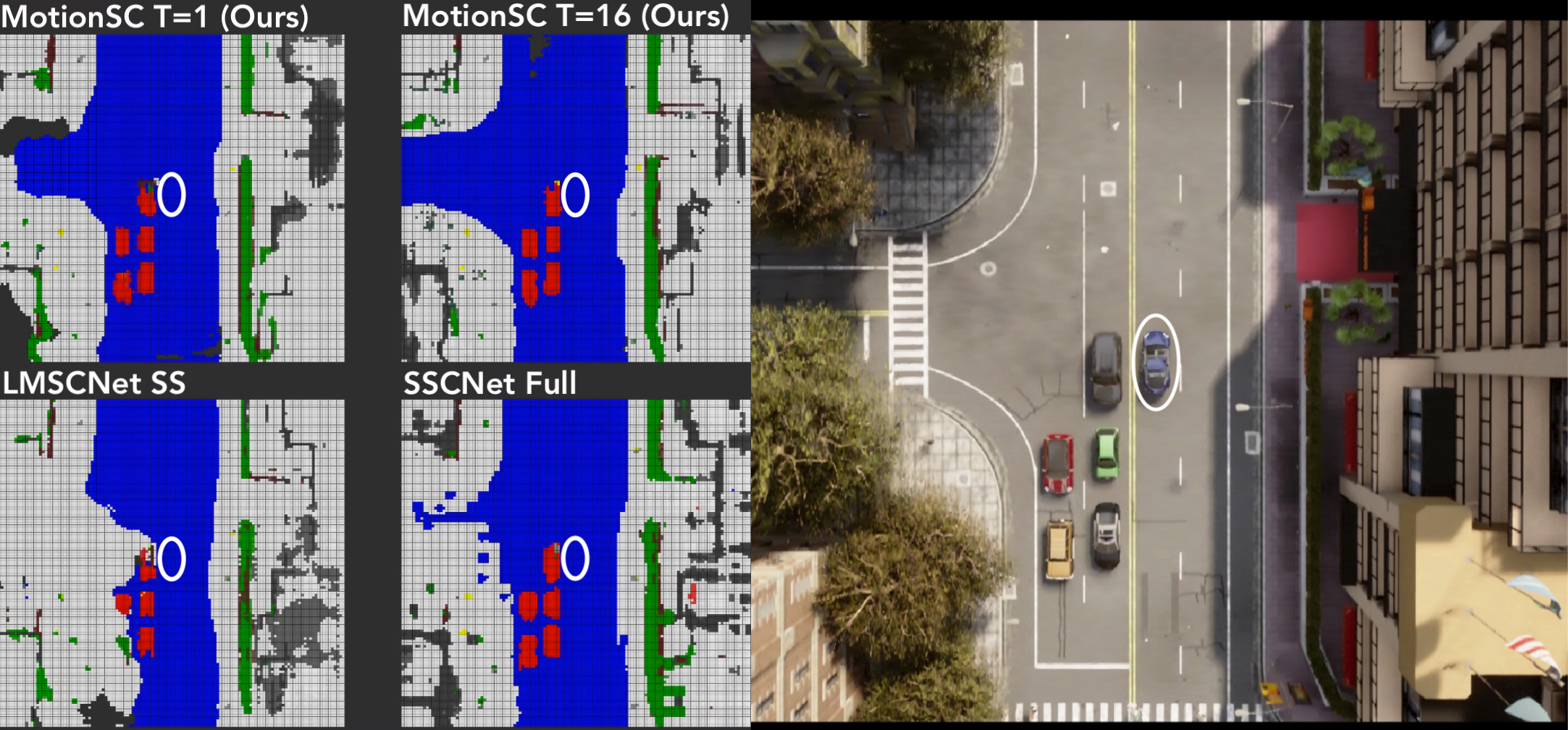}
    \caption{\textcolor{black}{Qualitative comparison of MotionSC at $T=1$ and $T=16$ with SSCNet and LMSCNet. The left four images show predictions from each network, visualized from a bird's eye view. The right image shows the ground truth bird's eye view image of the scene for validation. Note that the ego vehicle is not in the predicted scenes, and is circled in white. In this frame, the ego vehicle's view of the road to the left is occluded. LMSCNet, SSCNet, and MotionSC ($T=1$) are all susceptible to occlusions, and have noise in the output, although MotionSC has the most complete road. When $T=16$, MotionSC successfully incorporates past point clouds to complete the road.} Videos are included on our website.}
    \label{fig:Qualitative Comparison}
    \vspace{-4mm}
\end{figure}

As was shown quantitatively, MotionSC achieves a high level of geometric completeness and semantic segmentation due to temporal information. In Fig.~\ref{fig:Qualitative Comparison}, a scene from the test set of our data set is depicted, which shows the ego vehicle driving next to multiple cars. \textcolor{black}{The neighboring vehicle obscures the view of the nearby street, causing noise in both baselines and MotionSC with $T=1$. In contrast, MotionSC ($T=16$) is able to complete the scene from past scans, and accurately forms the road to the left of the scene. MotionSC ($T=1$) still outperforms the baselines, as was demonstrated quantitatively on both our data set and Semantic KITTI.}


Each network is able to generate a complete scene with semantic labels to some degree. SSCNet creates a smooth scene, although it is unable to complete the road to the left of the scene due to the occlusion. LMSCNet is most susceptible to noise due to occlusions, which is evident again in the street to the left and in individual vehicles. MotionSC in contrast, is able to generate the most smooth scene of the view-volume networks, with the road and sidewalk clearly labeled, and minimal occlusions.

%% file: conclusion.tex
\section{Conclusion}
\label{sec:conclusion}


This paper addressed a need for semantic scene completion data by creating a novel outdoor data set with accurate and complete dynamic scenes. The data set uses the CARLA simulator \cite{CARLA} with similar parameters to the real-life Semantic KITTI~\cite{KITTI} data set to promote generalization in future works. We trained a real-time dense local semantic mapping algorithm to enhance semantic scene completion using temporal information. Our network shows that the proposed data set can quantify and supervise accurate scene completion in the presence of dynamic objects, encouraging future developments of dynamic mapping algorithms by the community. 

Future work includes improving the MotionSC algorithm to develop a more robust real-time local semantic mapping algorithm by adding recurrency or quantifiable uncertainty. \textcolor{black}{Additional direction includes extending the proposed data construction method to the real world. While multiple sensors may not directly transfer to the real world, correct dynamic scenes may be formed by segmenting instances and using partial views to complete each actor, or importance sampling of point clouds from a multi-agent environment.}